\documentclass[letterpaper]{article} 
\usepackage[submission]{aaai23}  
\usepackage{times}  
\usepackage{helvet}  
\usepackage{courier}  
\usepackage[hyphens]{url}  
\usepackage{graphicx} 
\usepackage{natbib}  
\usepackage{caption} 
\usepackage{algorithm}
\usepackage{algorithmic}
                              
\usepackage[dvipsnames]{xcolor}
\usepackage{enumitem}
\usepackage{multirow}
\usepackage{graphicx}
\usepackage{amsthm}
\usepackage{amsmath}
\usepackage{amsfonts}
\usepackage{array}
\usepackage{color}
\usepackage{colortbl}
\usepackage{pifont}
\usepackage{utfsym}
\usepackage{subcaption}
\usepackage{textcomp}
\usepackage{lipsum}  
\usepackage{wasysym}
 
\usepackage{booktabs}
\usepackage{soul}

\definecolor{mygreen}{RGB}{30, 147, 116}

\definecolor{myred}{RGB}{216, 50, 40}
\definecolor{myblue}{RGB}{60, 93, 159}
\definecolor{myorange}{RGB}{132, 0, 164}
\usepackage{newfloat}
\usepackage{listings}
\DeclareCaptionStyle{ruled}{labelfont=normalfont,labelsep=colon,strut=off} 
\floatstyle{ruled}
\newfloat{listing}{tb}{lst}{}
\floatname{listing}{Listing}
\title{Exploring Sparsity in Graph Transformers}
\author {
   Chuang Liu,\textsuperscript{\rm 1}
   Yibing Zhan,\textsuperscript{\rm 2}
   Xueqi Ma,\textsuperscript{\rm 3}
   Liang Ding,\textsuperscript{\rm 2}
   Dapeng Tao,\textsuperscript{\rm 4}\\
   Jia Wu,\textsuperscript{\rm 5}
   Wenbin Hu,\textsuperscript{\rm 1 \thanks{Corresponding Author}}
   Bo Du,\textsuperscript{\rm 1 }
}
\affiliations {
    \textsuperscript{\rm 1} School of Computer Science, Wuhan University, Wuhan, China\\
    \textsuperscript{\rm 2} JD Explore Academy, JD.com, China\\
    \textsuperscript{\rm 3} School of Computing and Information Systems, The University of Melbourne, Melbourne, Australia \\
    \textsuperscript{\rm 4} School of Computer Science, Yunnan University, Kunming, China \\
    \textsuperscript{\rm 5} School of Computing, Macquarie University, Sydney, Australia\\
    \{chuangliu, hwb, dubo\}@whu.edu.cn,
zhanyibing@jd.com,
xueqim@student.unimelb.edu.au,
liangding.liam@gmail.com,
dptao@ynu.edu.cn,
jia.wu@mq.edu.au
}

\usepackage{bibentry}

\begin{document}

\maketitle

\begin{abstract}
Graph Transformers (GTs) have achieved impressive results on various graph-related tasks. However, the huge computational cost of GTs hinders their deployment and application, especially in resource-constrained environments. Therefore, in this paper, we explore the feasibility of sparsifying GTs, a significant yet under-explored topic. We first discuss the redundancy of GTs based on the characteristics of existing GT models, and then propose a comprehensive \textbf{G}raph \textbf{T}ransformer \textbf{SP}arsification (GTSP) framework that helps to reduce the computational complexity of GTs from four dimensions: the input graph data, attention heads, model layers, and model weights. Specifically, GTSP designs differentiable masks for each individual compressible component, enabling effective end-to-end pruning. We examine our GTSP through extensive experiments on prominent GTs, including GraphTrans, Graphormer, and GraphGPS. The experimental results substantiate that GTSP effectively cuts computational costs, accompanied by only marginal decreases in accuracy or, in some cases, even improvements. For instance, GTSP yields a reduction of 30\% in Floating Point Operations while contributing to a 1.8\% increase in Area Under the Curve accuracy on OGBG-HIV dataset. Furthermore, we provide several insights on the characteristics of attention heads and the behavior of attention mechanisms, all of which have immense potential to inspire future research endeavors in this domain. Code is available at \url{https://anonymous.4open.science/r/gtsp}. 
\end{abstract}

\section{Introduction}
\label{sec:introduction}

Recently, Graph Transformer (GT)~\cite{graph-transformer} and its variants~\cite{graphtrans,graphormer-v1,graphgps} have achieved performance comparable or superior to state-of-the-art Graph Neural Networks (GNNs) on a series of graph-related tasks, particularly on graph-level tasks such as graph classification. Nevertheless, GTs are more resource-intensive than GNNs due to their stack of multi-head self-attention modules (MHA), which suffer from quadratic complexity that renders their deployment impractical under resource-limited scenarios. Therefore, 
reducing the computational costs of GTs while maintaining their performance has become highly significant.

Graph model compression has experienced a surge of interest, with pruning emerging as a prominent technique~\cite{surve-graph-pruning}. Currently, several works~\cite{ugs,gebt,glt,cgp} have been proposed to prune the GNN models from the perspectives of the model weights, the graph adjacency matrix, and the feature channels. These pruning works can accelerate GNNs' training and inference for node classification tasks on large-scale graphs. However, it remains unclear whether these methods are still effective for GTs due to the following differences: \textbf{1)} GNN pruning methods mainly target node classification tasks and prune edges from graphs. In contrast, GT mainly focuses on graph classification tasks, meaning that the efficiency is influenced by the number of input nodes to a greater extent. \textbf{2)} The architectural designs of GNNs and GTs differ considerably. Therefore, new pruning methods specifically designed for GTs are required. However, no study has yet been conducted to explore sparsification techniques explicitly tailored for GTs. 

\begin{figure}[!t] 
\begin{center}
\includegraphics[width=0.9 \linewidth]{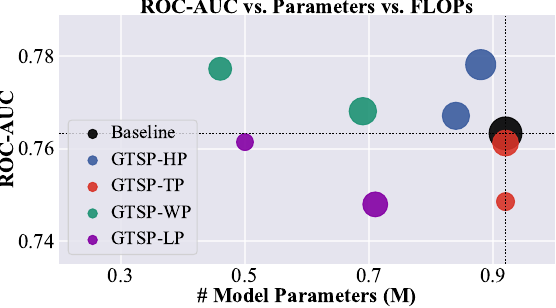}
\end{center}
\caption{Performance (\textit{y-axis}) analysis of GraphTrans baseline and our GTSP with varying model sizes  (\textit{x-axis}) and inference FLOPs (\textit{the size of markers}) on OGBG-HIV.  HP, TP, WP, and LP correspond to head pruning, token pruning, weight pruning, and layer pruning, respectively. Notably, GTSP achieves comparable or sometimes even slightly better performance than the baseline model with far fewer parameters and FLOPs.}
\label{fig:para}
\end{figure}

Therefore, in this paper, we aim to investigate the feasibility of sparsification techniques for GTs. To this end, we propose the \textbf{G}raph \textbf{T}ransformer \textbf{SP}arsification framework (GTSP), which is the first framework designed to reduce the computational resources of GTs. Our exploration begins by examining the redundancy present in GTs. In this context, redundancy refers to extraneous information that can be eliminated without significantly impacting the performance—a phenomenon commonly observed in transformer models~\cite{redundancy-ana,redundancy-attention}. We conduct a comprehensive analysis of GT redundancy across various dimensions, such as input nodes, attention heads, model layers, and model weights. Such an analysis provides valuable insights into the nature of GTs and guides our approach to effectively compress them. Subsequently, we propose a range of sparsification methods within the GTSP framework to discard the aforementioned  redundancy. Specifically, GTSP incorporates a learnable token selector to dynamically prune input nodes, customizes an importance score to guide the attention head pruning process, employs skip connectivity patterns across different GT layers, and dynamically extracts sparse sub-networks.  Benefiting from the above design features, GTSP effectively reduces the model redundancy, leading to reduced computational resource requirements without compromising on performance. Please note that this paper  focuses on  exploring the feasibility of pruning  four compressible components in GTs, and analyzing the advantages and insights of pruning each individual component.  A joint pruning of all compressible components in GTs in consideration of the complicated interactions between them is an area worth exploring in the future.



To evaluate the effectiveness of our GTSP, we conduct extensive experiments on various commonly-used datasets, including the large-scale Open Graph Benchmark~\cite{ogb-dataset}, with three prominent GTs: GraphTrans~\cite{graphtrans}, Graphormer~\cite{graphormer-v1}, and GraphGPS~\cite{graphgps}. The experimental results demonstrate that our GTSP reliably improves the efficiency of GT models while maintaining their performance. For example, in Figure~\ref{fig:para},  our GTSP-WP (\text{\Large \textcolor{mygreen}{$\bullet$}})  slightly  outperforms the baseline full models (GraphTrans) with only 50\% of the parameters and 69.8\% of FLOPs. 

\textbf{Insights:} Our thorough analysis of GTSP has yielded several noteworthy insights: \textbf{1)} Removing a large percentage of the attention heads does not significantly affect performance. The attention heads serve distinct roles in capturing various types of information, such as long-range and neighbor information within the graph. \textbf{2)} In general, up to 50\% of the model's neurons are redundant and can be pruned, which prevents over-fitting during training and can potentially improve accuracy. \textbf{3)}  Redundancy is evident among adjacent layers within the network, with deeper layers displaying even greater redundancy in relation to their neighboring layers. Selectively trimming certain layers not only accelerates training but also alleviates the over-smoothing issue on the graph. \textbf{4)}  Finally, we observe that GTs tend to prioritize important nodes while disregarding redundant nodes that can be safely removed from the graph.

\textbf{Contributions:} Our main contributions can be summarized as follows:  \textbf{1)} We present a premier investigation into the redundancy of GT models, which enhances the understanding of these models and provides valuable guidance on the design of sparsification methods. \textbf{2)} For the first time, we propose a comprehensive framework for sparsifying GT models, known as GTSP. This framework aims to improve the efficiency of GT models by sparsifying their components across four dimensions: input nodes, attention heads, model layers, and model weights. \textbf{3)} Experimental results on large-scale datasets with three popular GTs consistently validate the effectiveness and versatility of GTSP in offering enormous computation savings without compromising on accuracy. Additionally, we provide several valuable insights into the characteristics of existing GT models, which have the potential to inspire further research in this field.

\begin{figure*}[!t] 
\begin{center}
\includegraphics[width=0.95\linewidth]{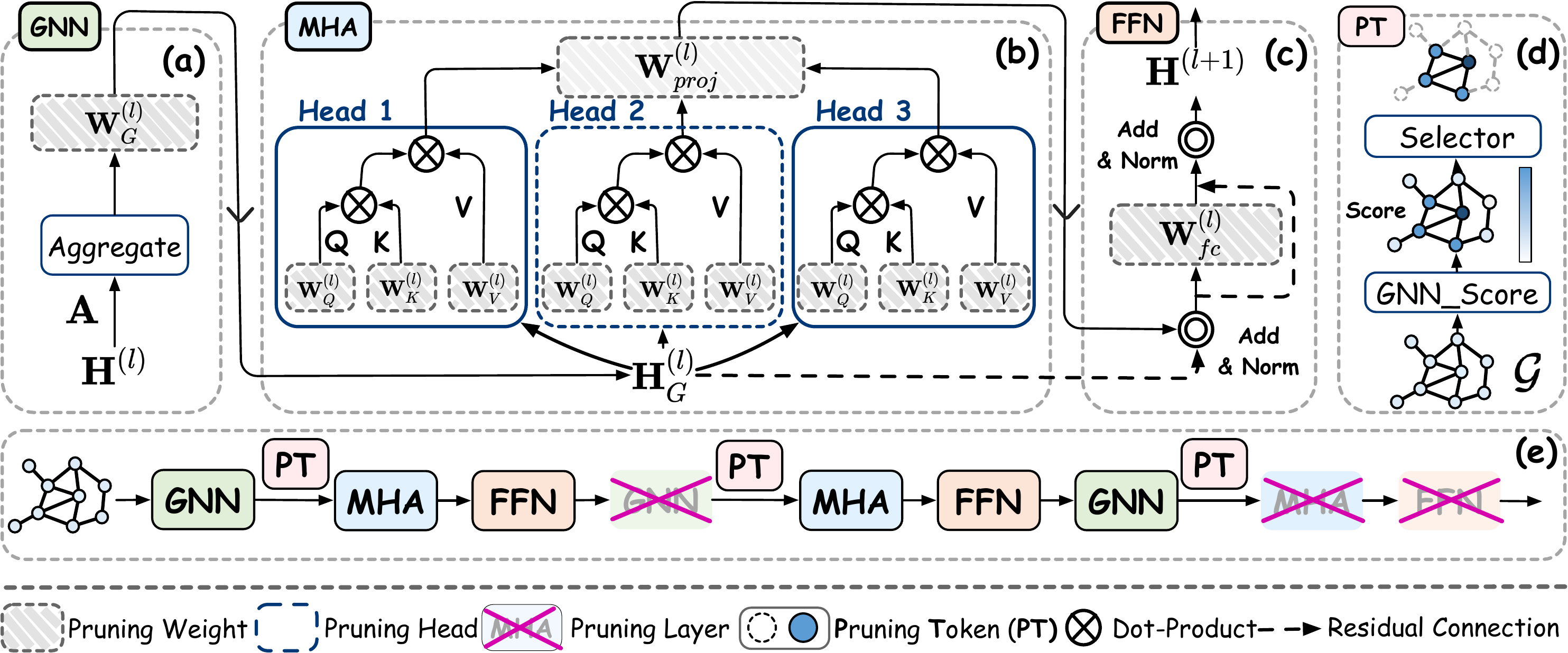}
\end{center}
\caption{Overview of our proposed graph transformer sparsification framework (GTSP). Note that each compressible component is pruned separately in this paper.}
\label{fig:model}
\end{figure*}

\section{Background}
\label{sec:related work}

\paragraph{Notations}  A graph $\mathcal{G}$ can be represented by an adjacency matrix $\mathbf{A} \in \{0, 1\}^{ n \times n}$ and a node feature matrix $\mathbf{X} \in \mathbb{R}^{ n \times d}$, where $n$ is the number of nodes,  $d$ is the dimension of node features, and  $\mathbf{A}[i, j]=1$ if there exists an edge between node $v_{i}$ and node $v_{j}$ (otherwise, $\mathbf{A}[i, j]=0$). 

\paragraph{Transformer} The vanilla Transformer~\cite{transformer} contains two key components: a multi-head self-attention (MHA) module and a position-wise feed-forward network (FFN).  Given the input matrix of node embeddings $\mathbf{H} \in \mathbb{R}^{n \times d}$, where $d$ represents the hidden dimension, a MHA module at layer $l$ is computed as follows:
\begin{equation}
\operatorname{MHA}(\mathbf{H}^{(l)})= \operatorname{Att}^{(l)}\left(\mathbf{W}_Q^{(l)}, \mathbf{W}_K^{(l)}, \mathbf{W}_V^{(l)}, \mathbf{H}^{(l)}\right), 
\label{eq:mha}
\end{equation}
where $\mathbf{W}_Q^{(l)}, \mathbf{W}_K^{(l)}, \mathbf{W}_V^{(l)} \in \mathbb{R}^{d \times d^{\prime}}$ denote the query, key, and value projection matrices, respectively. $\operatorname{Att}^{(l)}$ denotes the self-attention function and $d^{\prime}$ denotes the output dimension. Note that Eq.~(\ref{eq:mha}) denotes the single-head self-attention module, which can straightforwardly generalize to MHA. To construct a deeper model, each MHA layer and FFN layer is accompanied by a residual connection and subsequently normalized by means of layer normalization (LN).

\paragraph{Graph Transformer} Many transformer variants, inspired by transformer models, have been successfully applied to graph modeling. These variants often outperform or match GNNs across various tasks.  Unlike images and texts, graphs possess inherent structural characteristics; hence, the graph structure is crucial in graph-related tasks. Consequently, the most straightforward way to incorporate graph structure information is to combine GNNs with the transformer architecture~\cite{grover,graphtrans,graphgps}. This integration can be represented as follows:
\begin{equation*}
\mathbf{H}_{\text{G}}^{(l+1)}=\operatorname{GNN}^{(l)}\left(\mathbf{H}^{(l)}, \mathbf{A}\right), 
\mathbf{H}^{(l+1)}=\operatorname{MHA}^{(l)}\left(\mathbf{H}_{\text{G}}^{(l+1)}\right),
\end{equation*}
where $\operatorname{GNN}^{(l)}$ denotes a GNN layer. Additionally, several existing works have attempted to compress the graph structure into positional embedding (PE) vectors that are then incorporated into the input features~\cite{graph-transformer,san,egt}. Alternatively, the graph structure can be injected into the attention computation through bias terms~\cite{graphormer-v1}. For further information on these topics, refer to recent reviews of GTs~\cite{transformer-review}. However, most of these methods encounter challenges due to the time and memory constraints imposed by their complex frameworks, particularly the high computational complexity of MHA~\cite{ans-gt,nagphormer}. It is, therefore, crucial to identify approaches for reducing the computational resources required by GTs while preserving their performance.

\paragraph{Model Pruning}  Pruning is a promising method for reducing the memory footprint and computational cost by removing unimportant elements based on pre-defined scores~\cite{survey-pruning}. Various pruning methods, such as the magnitude-based, first-order and second-order based, and lottery ticket hypothesis-based methods~\cite{granet,growth-set,lth}, have been used to remove redundant weights. Additionally, some studies have explored the pruning of input tokens~\cite{ltp,evit}, attention heads~\cite{head-analysis,head-sixteen}, and even entire layers within the model architecture~\cite{layer-prune,wdpruning}. In the field of graphs, attempts have been made to co-prune model weights, graph adjacency matrices, and feature channels in order to speed up the training and inference of GNNs on large-scale graphs~\cite{gebt,ugs,glt,cgp}. However, to the best of our knowledge, no investigation has yet been conducted into the pruning of GTs.


\section{Methodology}
\label{sec:method}

This section explores redundancy within GT architectures and presents a comprehensive framework, called GTSP, which aims to enhance the efficiency of GTs through pruning. Specifically, GTSP is a mask-based pruning method, which contains three crucial steps in the overall pruning procedure:  \text{ \large \ding{202}}initializing masks for compressible components; \text{ \large \ding{203}}determining the values of these masks; and \text{ \large \ding{204}}pruning based on the masks. Figure~\ref{fig:model} illustrates the application of GTSP in compressing GTs across four dimensions: input data (\textbf{$\S$\ref{sec:pruning-token}}), attention heads (\textbf{$\S$\ref{sec:pruning-head}}), model layers (\textbf{$\S$\ref{sec:pruning-layer}}), and model weights (\textbf{$\S$\ref{sec:pruning-weight}}). Please note that this paper primarily focuses on the individual pruning of each component; the joint pruning of all components, considering their intricate interactions, is left as a topic for future research.

\begin{figure}[!t] 
\begin{center}
\includegraphics[width=1.0\linewidth]{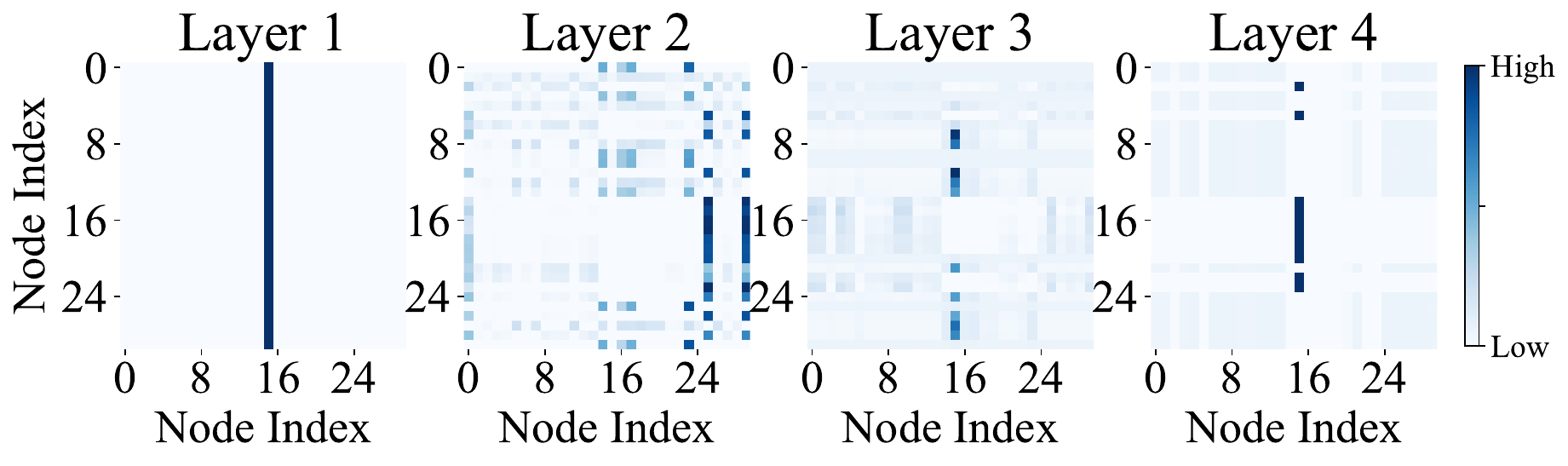}
\end{center}
\caption{\textbf{Token Redundancy.} Attention probability between query (\textit{y-axis}) and key (\textit{x-axis}) vectors obtained from GraphGPS. The probabilities in each row all sum to 1.}
\label{fig:redun-token}
\end{figure}

\subsection{Pruning Tokens}
\label{sec:pruning-token}

Self-attention is capable of modeling long-term dependencies. However, the computational complexity of computing the attention matrix increases quadratically with the length of the input tokens ($\mathcal{O}(n^{2})$). Consequently, when working with large graphs and limited resources, the attention operation becomes a bottleneck. To achieve efficiency improvements, we aim to eliminate less important or relevant tokens before they pass through the transformer layers. By reducing the number of tokens $n$ for subsequent blocks, we can reduce the complexity of both the MHA and FFN layers.




\paragraph{Analyzing Token Redundancy} 

Figure~\ref{fig:redun-token} illustrates the attention probability, which measures how much all other tokens attend to a specific token. It is worth noting that only a limited number of tokens (\textit{e.g.}, node 15) receive high attention scores, while others are considered less important and may be pruned. Furthermore, we offer supplementary validation in \textbf{$\S$\ref{sec:broader}} to support our findings.



\paragraph{Discarding Token Redundancy}

In general, we use a trainable mask to discard tokens with low importance scores for token sparsification in an end-to-end optimization. To accomplish this, \text{\large \ding{202}}we first initialize a binary decision mask  $\mathbf{M}_{T}^{(l)} \in \{0, 1\}^{n}$ that indicates whether a token should be dropped or kept. We then design a learnable prediction module that generates important scores for input nodes, \text{\large \ding{203}}which helps determine the values in the mask $\mathbf{M}_{T}^{(l)}$. Specifically, we project the tokens using a GCN~\cite{gcn} to capture both their feature and structure information:
\begin{equation}
\mathbf{S}_{T}^{(l)}  = \text{GCN} (\mathbf{A}, \mathbf{H}^{(l)}) \in \mathbb{R}^{n \times c},
\end{equation}
where $c$ is the dimension of the scores. To preserve the significant tokens and remove the useless ones, we first rank tokens in order of their scores and then prune them using a top-$k$ selection strategy. If the score $\mathbf{S}_{T}^{(l,i)}$ of token $i$ is smaller than the $k$ largest values among all tokens, $\mathbf{M}^{(l,i)}_{T}= 0$, which indicates that the token $i$ is pruned at layer $l$. 

However, performing the above operation is not straightforward in practice because using the top-$k$ operation to generate a mask is not differentiable, which hinders end-to-end training. To address this issue, we introduce the Gumbel-Softmax and straight-through tricks, which facilitate gradient back-propagation through the top-$k$ selection. Another obstacle arises when using GCN to generate scores. GNNs tend to share similar information among directly connected nodes. As a result, models may assign similar scores to nearby nodes with similar keys; this causes models to get stuck in significant local structures and select redundant nodes, while ignoring important nodes from other substructures and losing structure information. To address this issue, we propose applying perturbations to the significance scores. Therefore, the decision mask is calculated as follows:
\begin{equation}
\mathbf{M}_{T}^{(l)}  = \text{Gumbel-Softmax}(\mathbf{I}_{\lceil {p_s}\times {n}\rceil} \mathbf{S}_{T}^{(l)}) \in \{0, 1\}^{n},
\end{equation}
where $ \mathbf{I}_{\lceil {p_s}\times {n}\rceil} \in \displaystyle \mathbb{R}^{ n \times n}$ is a matrix generated by randomly dropping $\lceil {p_s}\times {n} \rceil$ non-zero elements of a unit matrix with $n$ dimensions, $\lceil \cdot \rceil$ is the rounding up operation,  and  $p_s$ is the score dropping rate. Finally, \text{\large \ding{204}}we prune tokens by applying the mask $\mathbf{M}_{T}^{(l)}$ to the node embedding matrix $\mathbf{H}^{(l)}$, and the mask is updated at each epoch.


\subsection{Pruning Attention Heads}
\label{sec:pruning-head}

The attention mechanism employed by current GTs incorporates multiple attention heads, commonly utilizing either four or eight self-attention heads. However, there is a dearth of analysis and discussion on the necessity of using such a substantial number of heads, as well as the specific information focused on by each head during the process of representation learning and its relevance to downstream tasks.

\begin{figure}[!t]
\centering
\begin{subfigure}{0.31\textwidth}
\includegraphics[width=0.95\linewidth]{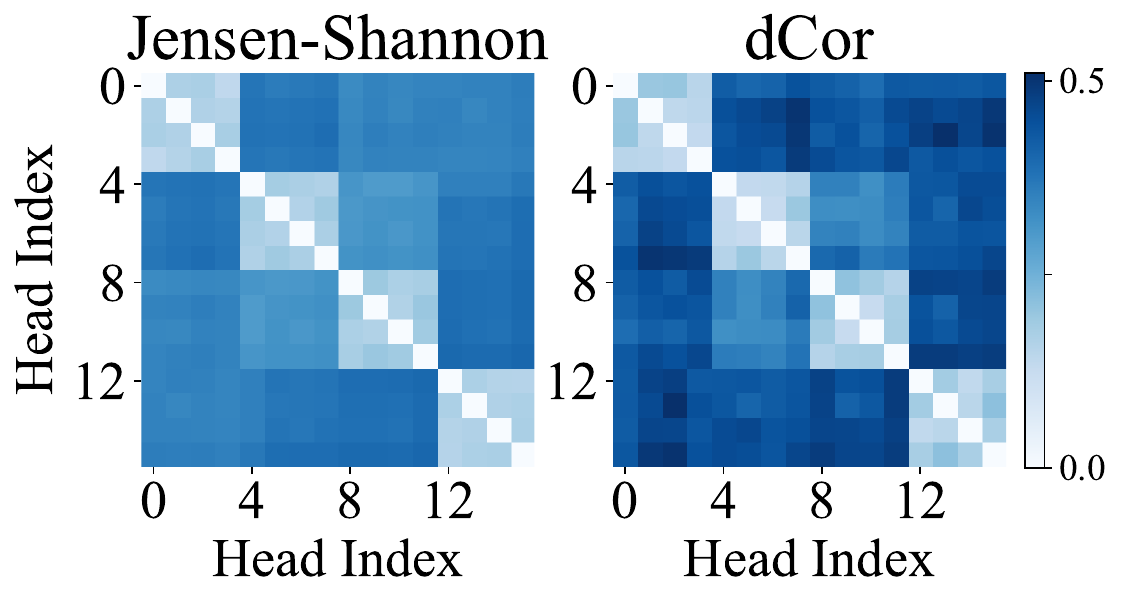} 
\caption{ {\footnotesize {Head Redundancy.}}}
\label{fig:redun-head}
\end{subfigure}
\hspace{-3mm}
\begin{subfigure}{0.16\textwidth}
\includegraphics[width=1.04\linewidth]{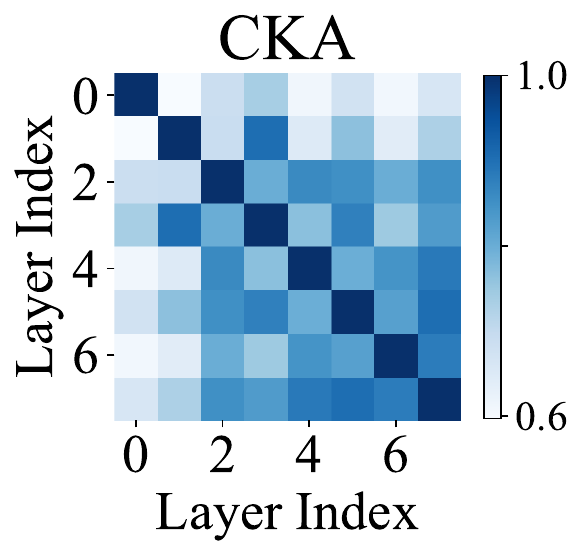}
\caption{  {\footnotesize Layer Redundancy} }
\label{fig:redun-layer}
\end{subfigure}
\caption{\textbf{(a) Head Redundancy.} Distance between different attention heads in GraphTrans (4-layer-4-head self-attention) is assessed using both the Jensen–Shannon and dCor metrics. Smaller distance values reflect more redundancy. \textbf{(b) Layer Redundancy.}  Pairwise similarity between the layers in GraphGPS is measured using the CKA metric. Darker colors represent higher levels of similarity.}
\label{fig:hybrid}
\end{figure}

\paragraph{Analyzing Head Redundancy}
To determine if the attention mechanism involves redundant computations, we calculate the similarity of head distributions. We define the attention redundancy matrix as the pairwise distance matrix of the $4 \times 4$ attention matrices from the GT model, using both node-level (Jensen–Shannon distance~\cite{JS-distance}) and graph-level (Distance Correlation, or dCor~\cite{dcor}) measures. Taking the GraphTrans model as an example, Figure~\ref{fig:redun-head} illustrates the redundancy (distance) among $16 \times 16$ pairs of attention matrices in a 4-layer-4-head configuration. It is evident that redundancy exists in the attention layers, particularly in the earlier ones. Additionally, the redundancy patterns remain consistent between two different distance measures used in our analysis. Based on these findings, we can conclude that certain attention heads can be safely removed.

\paragraph{Discarding Head Redundancy} 

In the initial step, \text{\large \ding{202}}we proceed with the initialization of a decision mask, denoted as $\mathbf{M}_{H}^{(l)} \in {0, 1}^{N_h}$, where $N_h$ represents the number of attention heads. Subsequently, \text{\large \ding{203}}we introduce a criterion for use in estimating the significance of each head. More precisely, the significance of a given head is determined based on the model's sensitivity to the masked head~\cite{head-sixteen,svite}. To calculate this significance value for each head, we use the following expression after applying the chain rule:
\begin{equation}
\mathbf{S}_H^{(l, h)}=\left|(\mathbf{Z}^{(l, h)})^{\mathrm{T}} \cdot \frac{\partial \mathcal{L}\left(\mathbf{H}^{(l)}\right)}{\partial \mathbf{Z}^{(l, h)}}\right|,
\end{equation}
where $\mathbf{Z}_{(l, h)}$ denotes the embeddings of the $l$-th layer and $h$-th head computed by Eq.~(\ref{eq:mha}). Additionally, $\mathcal {L}(\cdot)$ refers to the cross-entropy loss used in GTs. Once we have obtained importance scores for attention heads, we remove those with the smallest $\mathbf{S}_H^{(l, h)}$. \text{\large \ding{204}}To prune these heads, we modify the formula for $\operatorname{MHA}$ in Eq.~(\ref{eq:mha}):
\begin{equation*}
\operatorname{MHA}(\mathbf{H}^{(l)})=\sum_{h=1}^{N_h} \mathbf{M}_{H}^{(l,h)} \operatorname{Att}^{(l)}\left(\mathbf{W}_Q^{(l,h)}, \mathbf{W}_K^{(l)}, \mathbf{W}_V^{(l,h)}, \mathbf{H}^{(l)}\right).
\end{equation*} 

In addition, to prevent the premature removal of valuable attention heads, we propose a mechanism inspired by the concept of regrowth in weight pruning~\cite{cgp}. This mechanism utilizes magnitude gradients, $\left\|\partial \mathcal{L}\left(\mathbf{H}^{(l)}\right) /\partial \mathbf{Z}^{(l, h)}\right\|_{\ell_1}$, as a criterion for determining which (if any) pruned heads should be regenerated. The attention heads with the highest gradients will be reactivated.

\subsection{Pruning Layers}
\label{sec:pruning-layer}
GTs often combine GNN layers with transformer layers. For example, in GraphTrans, the initial layers are GNN layers followed by multiple transformer layers. In GraphGPS, there is one GNN layer followed by one transformer layer. Although different studies have proposed various combinations, none has investigated whether it is necessary to stack multiple layers or if redundancy exists between them.

\paragraph{Analyzing Layer Redundancy} 
To evaluate layer redundancy, we compare representations from different layers using linear Center Kernel Alignment (CKA)~\cite{cka}, which is a widely used method for identifying relationships between layers across architectures. In our analysis, we compute pairwise similarity between representations obtained from all eight intermediate layers (four GNN layers and four transformer layers) of GraphGPS and present the corresponding heatmaps in Figure~\ref{fig:redun-layer}. From the results, we can make several observations: \textbf{1)} Adjacent layers show notable similarity, indicating some redundancy among them.   \textbf{2)} Deeper layers exhibit greater similarity with each other, suggesting increased redundancy at deeper levels. This implies that these deeper layers may contribute minimally to the discriminative power of the model.  These observations justify the motivation to drop certain layers.  Furthermore,  GT models have uniform internal structures with identical input/output sizes for MHA, FFN, and GNN models. This uniformity allows us to remove any of these components while directly concatenating the remaining ones without causing any feature dimension compatibility issues.



\paragraph{Discarding Layer Redundancy} 
Building upon the previous study~\cite{layer-drop}, we simultaneously prune redundant GNN and transformer layers in GTs using a random drop-layer approach. For instance, let us consider pruning a MHA layer. The original formulation of the residual connection in GTs is as follows:
\begin{equation}
\mathbf{H}^{(l+1)}=\operatorname{LN}\left(\operatorname{MHA}\left(\mathbf{H}^{(l)}\right)\right)+\mathbf{H}^{(l)}.
\label{eq:residual}
\end{equation}

To determine whether to keep or drop a layer, \text{\large \ding{202}}we introduce a mask $\mathbf{M}^{(l)}_{L} \in \{0,1\}^{L}$, where $L$ is the number of layers, into the above equation. Consequently, Eq.~(\ref{eq:residual}) is reformulated as follows:
\begin{equation}
\mathbf{H}^{(l+1)}=\operatorname{LN}\left( \mathbf{M}^{(l)}_{L} \operatorname{MHA}\left(\mathbf{H}^{(l)}\right)\right)+\mathbf{H}^{(l)},
\label{eq:residual-mask}
\end{equation}
\text{\large \ding{203}}where the value of masks is sampled from a Bernoulli distribution. Specifically, if $\mathbf{M}^{(l)}_{L}=0$, \text{\large \ding{204}}it means that the layer is pruned and the node's $(l+1)$-th layer representation remains consistent with its $l$-th layer representation. On the other hand, if $\mathbf{M}^{(l)}_{L}=1$, then the node's $(l+1)$-th layer representation is updated.

\begin{table*}[!t]
\centering
\renewcommand\arraystretch{1.3} 
\setlength\tabcolsep{2.2pt} 
\caption{Performance measured by the number of \textbf{Para}meters / \textbf{F}LOPs \textbf{S}aving / \textbf{Acc}uracy ($\uparrow$) or\textbf{ ROC-AUC} ($\uparrow$). The results with higher accuracy than the baseline have been highlighted in \textbf{bold}. ``GraphTrans-S'' refers to the reduced-scale GraphTrans model, featuring a decreased number of layers. ``GNN-Dense" denotes the models solely composed of GNN layers. }
\label{tab:results}
\resizebox{0.9\textwidth}{!}{%
\begin{tabular}{@{}l|c|ccc|ccc|cccc@{}}
\toprule
\multirow{2}{*}{Models} & \multirow{2}{*}{Spar.} & \multicolumn{3}{c|}{\textbf{NCI1}}                                & \multicolumn{3}{c|}{\textbf{OGBG-HIV}}  & \multicolumn{4}{c}{\textbf{OGBG-Molpcba}}                                  \\ \cmidrule(l){3-12} 
                        &                            &\# \textbf{Para.} & \textbf{FS}       & \textbf{Acc.} (\%)       & \# \textbf{Para.} & \textbf{FS}      & \textbf{ROC-AUC}  & \# \textbf{Para.} & \textbf{FS}      & \textbf{ROC-AUC}(V) & \textbf{ROC-AUC}(T)     \\ \midrule
\rowcolor[HTML]{EFEFEF} \textbf{GraphTrans}              & 0 \%                      & 0.82M    & 0\%            & $83.11_{ \pm 1.78}$   & 0.92M    & 0\%          & $0.7633_{ \pm 0.0111}$ & 2.41M    & 0\%          & $0.2893_{ \pm 0.0050}$ & $0.2756_{ \pm 0.0039}$   \\ \midrule
\text{\Large \textcolor{myblue}{$\bullet$}} \textbf{GTSP-HP}                  & 25\%                      & 0.75M        & 6.1\%  & $82.62_{ \pm 1.41}$ & 0.88M         & 8.52\% &  $ \textbf{0.7782}_{ \pm \textbf{0.0064}}$ & 2.34M    & 4.4\%          & $0.2871_{ \pm 0.0061}$ & $ \textbf{0.2756}_{ \pm \textbf{0.0075}}$ \\
\text{\Large \textcolor{myred}{$\bullet$}} \textbf{GTSP-TP}                  & 25\%                      & 0.82M         &  22.8\%                 & $82.59_{ \pm 1.47}$ &  0.92M        &  21.5\%                & $0.7612_{ \pm 0.0116}$ & 2.41M    & 20.7\%          & $0.2604_{ \pm 0.0068}$ & $0.2451_{ \pm 0.0060}$ \\
\text{\Large \textcolor{mygreen}{$\bullet$}} \textbf{GTSP-WP}                  & 25\%                      & 0.61M         & 18.7\%                  & $82.24 _{ \pm 1.60 }$ &  0.69M        &  16.5\%               & $\textbf{0.7681}_{ \pm \textbf{0.0225}}$  & 1.80M    & 20.3\%          & $\textbf{0.2893}_{ \mathbf{\pm} \textbf{0.0012}}$ & $\textbf{0.2794}_{ \pm \textbf{0.0025}}$ \\
\text{\Large \textcolor{myorange}{$\bullet$}} \textbf{GTSP-LP}                  & 25\%                      & 0.64M         & 23.5\%                & $82.53_{ \pm 1.71}$ &   0.71M       &   24.3\%              & $0.7479_{ \pm 0.0360}$ & 1.90M    & 22.1\%          & $0.2863_{ \pm 0.0019}$ & $0.2749_{ \pm 0.0015}$ \\ \midrule
\text{\Large \textcolor{myblue}{$\bullet$}}  \textbf{GTSP-HP}                  & 50\%                      & 0.69M          &   12.2\%              & $82.23_{ \pm 1.67}$ &   0.84M       & 17.1\%                 & $\textbf{0.7671}_{ \pm \textbf{0.0158}}$ & 2.27M    & 8.9\%          & $0.2712_{ \pm 0.0081}$ & $0.2615_{ \pm 0.0064}$ \\
\text{\Large \textcolor{myred}{$\bullet$}} \textbf{GTSP-TP}                  & 50\%                      &  0.82M        &   47.4\%               & $82.77_{ \pm 1.79}$ &   0.92M       & 45.3\%             & $0.7485_{ \pm 0.0103}$ & 2.41M    & 39.2\%          & $0.2573_{ \pm 0.0073}$ & $0.2436_{ \pm 0.0074}$ \\
\text{\Large \textcolor{mygreen}{$\bullet$}} \textbf{GTSP-WP}                  & 50\%                      &  0.41M        & 37.4\%                & $81.91 _{ \pm 2.06}$ & 0.46M         & 30.2\%                & $\textbf{0.7773}_{ \pm \textbf{0.0073}}$ & 1.20M    & 40.6\%          & $0.2868_{ \pm 0.0010}$ & $\textbf{0.2774}_{ \pm \textbf{0.0033}}$ \\
\text{\Large \textcolor{myorange}{$\bullet$}} \textbf{GTSP-LP}                 & 50\%                      & 0.44M       &  47.1\%               & $82.46_{ \pm 1.93}$ &  0.50M       & 48.8\%             & $0.7614_{ \pm 0.0176}$ & 1.41M    & 46.6\%          & $0.2748_{ \pm 0.0013}$ & $0.2664_{ \pm 0.0034}$  \\ \midrule
 \textbf{GraphTrans-S}        & 0\%                       &  0.56M         &  47.1\%              & $81.97_{ \pm 1.98}$ &   0.51M       &   48.8\%                & $0.7617_{ \pm 0.0176}$ & 2.00M    & 46.6\%          & $0.2830_{ \pm 0.0034}$ & $0.2752_{ \pm 0.0043}$ \\
 \textbf{GNN-Dense }              & 0\%                       & 0.58M         &  88.3\%              & $80.00_{ \pm 1.40}$ & 0.18M        &   95.4\%              & $0.7575_{ \pm 0.0104}$ & 1.60M    & 43.1\%          & $0.2305_{ \pm 0.0027}$ & $0.2266_{ \pm 0.0028}$ \\ \bottomrule
\end{tabular}%
}
\end{table*}

\subsection{Pruning Weights}
\label{sec:pruning-weight}

Over-parameterization is a common issue in neural networks that causes further information redundancy. To tackle this issue in GTs, we draw inspiration from sparse training techniques that dynamically extract and train sparse sub-networks instead of the entire model~\cite{survey-pruning}. 


\paragraph{Discarding Weight Redundancy} 
Generally speaking, we mask small-magnitude model weights in our approach. Specifically, during the model initialization stage, \text{\large \ding{202}}we create non-differentiable binary masks $\mathbf{M}_{W}$ that match the size of the model weights in different layers, $\boldsymbol{W}$. Initially, all elements in the mask are set to $1$. At regular intervals, \text{\large \ding{203}}our pruning strategy updates the mask matrix by setting parameters below a threshold to $0$. \text{\large \ding{204}}The weight matrix is then multiplied by this updated mask to determine which weights participate in the subsequent forward execution of the graph. This procedure can be described as follows:
\begin{equation}
\begin{aligned}
&\mathrm{idx}=\operatorname{TopK}(-|\mathbf{W}|,\lceil p_{w} \|\mathbf{W} \|_{0} \rceil); \\& \mathbf{M}_{W}^{\prime} = \operatorname{Zero}(\mathbf{M}_{W}, \mathrm{idx}); \mathbf{W}^{\prime} = \mathbf{M}_{W}^{\prime} \odot \mathbf{W}, 
\label{eq:prune-w}
\end{aligned}
\end{equation}
where $\operatorname{TopK}$ is the function that returns the indices of the top $\lceil p_{w} \|\mathbf{W} \|_{0}\rceil$ values in $|\mathbf{W}|$, $\operatorname{Zero}$ is the function that sets the values in $\mathbf{M}_{W}$ with indices $\mathrm{idx}$ to $0$,  $\mathbf{W}^{\prime}$ is the pruned weight matrix,  $p_{w}$ is the sparsity of the model weights, $\|\mathbf{W} \|_{0}$ is the number of model weights, and $\odot$ is the element-wise product. We utilize gradual magnitude pruning, as demonstrated in previous studies~\cite{gradual-prune,granet} to gradually prune the weights over $m$ iterations until the desired sparsity is reached. If we perform sparsification of all elements every $\Delta t$ steps, we can determine the pruning rate at each iteration $t$ as follows:
\begin{equation}
\begin{aligned} 
p_{(w,t)}=p_{f}+\left(p_{i}-p_{f}\right)&\left(1-\frac{t-t_{0}}{m \Delta t}\right)^{3}, 
\label{eq:gradual-pruning}
\end{aligned}
\end{equation}
where $p_{i}/p_{f}$ is the initial/target sparsity degree, and $t_{0}$ is the epoch at which gradual pruning begins. The pruning scheme described above involves initially pruning a large number of redundant connections, followed by gradually reducing the number of connections pruned as fewer remain.

In addition, premature pruning may occur during the pruning process, particularly in early iterations, resulting in the loss of important information. To address this issue, we propose incorporating the gradient-based regrowth schemes~\cite{growth-rigl} into the gradual pruning schedule. The regrowth operation, which is performed at regular intervals ($\Delta t$) throughout training, serves to reactivate weights with high-magnitude gradients. It is worth noting that previous studies have demonstrated the efficiency of this regrowth scheme for training~\cite{granet,cgp}.

\subsection{Complexity Analysis}

The computational costs of the GNN, MHA, and FFN modules in GTs are $\mathcal{O}\left(\|\boldsymbol{A}\|_0 d+ n d^2\right)$, $\mathcal{O}(n^{2}d)$, and $\mathcal{O}(nd^{2})$, respectively. Our GTSP aims to reduce the computational complexity of these modules. The additional computation introduced by GTSP primarily arises from the element-wise product between weights and masks, which is acceptable. For a detailed discussion on the complexity and FLOPs, please refer to the Appendix.

\section{Experiments}
\label{sec:experiment}

In this section, we validate the effectiveness of our proposed GTSP on three benchmark graph datasets, including OGB datasets. We demonstrate that GTSP is efficient and accurate (\textbf{$\S$\ref{sec:acc}}) and can generalize well to various baseline GT models  (\textbf{$\S$\ref{sec:general}}). Furthermore, we present insightful findings from our GTSP research (\textbf{$\S$\ref{sec:broader}}), such as a deeper understanding of attention heads and the advantage of using GTSP to mitigate over-fitting and over-smoothing issues.

\subsection{Experimental Settings}

\paragraph{Datasets} We choose three graph classification benchmarks: one small dataset (NCI1) and two large-scale datasets from Open Graph Benchmark (OGBG-HIV and OGBG-Molpcba)~\cite{ogb-dataset}. These datasets consist of approximately $4,000$, $41,127$, and $437,929$ graphs, respectively. We strictly follow the original settings of these datasets, including their splitting methods.



\paragraph{Implementation Details} 
We evaluate the effectiveness of our GTSP on three commonly-used GT models: GraphTrans~\cite{graphtrans}, GraphGPS~\cite{graphgps}, and Graphormer~\cite{graphormer-v1} by parameters, FLOPs, and a graph classification metric (accuracy or ROC-AUC). We aim to maintain the original settings of these models 
as much as possible. Our main results are based on 10 runs, except for OGBG-Molpcba, which is based on five runs. All models are trained using NVIDIA A100 GPUs (40G). The detailed parameter settings can be found in the Appendix.


\begin{table}[!t]
\centering
\renewcommand\arraystretch{1.2} 
\setlength\tabcolsep{2pt} 
\caption{Performance of other Graph Transformer models on the NCI1 dataset measured by the number of \textbf{Para}meters / \textbf{F}LOPs \textbf{S}aving / \textbf{Acc}uracy ($\uparrow$). The results with higher accuracy than the baseline have been highlighted in \textbf{bold}.}
\label{tab:other-gt}
\resizebox{0.48\textwidth}{!}{%
\begin{tabular}{@{}l|c|ccc|ccc@{}}
\toprule
                         &                                               & \multicolumn{3}{c|}{\textbf{GraphGPS}} & \multicolumn{3}{c}{\textbf{Graphormer}}                                                                                                                                                            \\ \cmidrule(l){3-5} \cmidrule(l){6-8} 
\multirow{-2}{*}{Models} & \multirow{-2}{*}{Spar.}                    & \textbf{\# Para.}                                     & \textbf{FS}                                                              & \textbf{Acc.} (\%)  & \textbf{\# Para.}                                     & \textbf{FS}                                                              & \textbf{Acc.} (\%)                                         \\ \midrule
\rowcolor[HTML]{EFEFEF} 
\textbf{Base}               & 0 \%                                          & 0.18M                                         & 0\%                                          & $83.71_{ \pm 1.77}$     & 0.17M                                         & 0\%                                          & $83.36_{ \pm 1.18}$                     \\ \midrule
\text{\Large \textcolor{myblue}{$\bullet$}} \textbf{HP}                   & 25\%                                          & 0.16M                                             & 12.0\%                                                        &   $\textbf{84.46}_{ \pm \textbf{1.58}}$    &  0.16M                                            &  12.0\%                                                                       &         $82.07_{ \pm 1.98}$                                       \\
\text{\Large \textcolor{myred}{$\bullet$}} \textbf{TP}                  & 25\%                                          & 0.18M                                             & 22.9\%                                                                         &   $83.24_{ \pm 1.20}$   & 0.17M                                            &  22.9\%                                                                     &    $82.47_{ \pm 1.96}$                                            \\
\text{\Large \textcolor{mygreen}{$\bullet$}} \textbf{WP}                   & 25\%                                          & 0.13M                                              & 20.9\%                                                                        &     $\textbf{83.72}_{ \pm \textbf{1.81}}$       & 0.13M                                             & 20.9\%                                                                        &  $\textbf{83.57}_{ \pm \textbf{1.11}}$                                   \\
\text{\Large\textcolor{myorange}{$\bullet$}} \textbf{LP}                   & 25\%                                          & 0.15M                                             & 22.4\%                                                                        &          $83.52_{ \pm 1.00}$   & 0.15M                                           &   22.4\%                                                                      & $\textbf{83.52}_{ \pm \textbf{1.91}}$                                    \\ \midrule
\text{\Large\textcolor{myblue}{$\bullet$}} \textbf{HP}                   & 50\%                                          &  0.14M                                            & 23.6\%                                                       &   $83.02_{ \pm 2.19}$   & 0.14M                                             &  23.6\%                                                                      & $82.46_{ \pm 2.42}$                                        \\
\text{\Large \textcolor{myred}{$\bullet$}} \textbf{TP}                   & 50\%                                          &  0.18M                                            & 40.0\%                                                                       &         $82.92_{ \pm 1.60}$ & 0.17M                                             &   40.0\%                                                                     &   $82.77_{ \pm 1.86}$                                                                           \\
\text{\Large\textcolor{mygreen}{$\bullet$}} \textbf{WP}                   & 50\%                                          &  0.09M                                            &  45.7\%                                                                      &         $\textbf{83.82}_{ \pm \textbf{1.58}}$        &  0.09M                                            &   45.7\%                                                                      &   $\textbf{83.48}_{ \pm \textbf{1.20}}$                                                                     \\
\text{\Large\textcolor{myorange}{$\bullet$}} \textbf{LP}                   & 50\%                                          & 0.11M                                             & 47.1\%                                                                       &        $82.80_{ \pm 1.22}$  & 0.11M                                           &     47.1\%                                                                     &  $83.26_{ \pm 1.34}$                                                                              
 \\\bottomrule
\end{tabular}%
}
\end{table}


\subsection{Accuracy w.r.t Efficiency} 
\label{sec:acc}
We evaluate the effectiveness of  GTSP based on parameters, FLOPs, and accuracy. The comparison between GTSP and  baseline models (including reduced-scale GraphTrans and GNN-Dense) is presented in Table~\ref{tab:results}. Based on these results, we can make several inspiring observations: \textbf{1)} Our GTSP achieves better accuracy and computation trade-offs than baseline models. \textbf{2)} As for pruning heads, GTSP can reduce the number of heads by $50\%$ while still achieving a $0.5\%$ improvement in accuracy, which indicates that GTSP can successfully reduce the head redundancy.  \textbf{3)} Token pruning does tend to significantly reduce FLOPs (\textit{e.g.}, $47.4\%$ FLOPS reduction on NCI1), but it also leads to a relatively larger accuracy degradation probably because of the relatively small number of nodes in these datasets (around $20–30$ nodes in a graph). \textbf{4)} Pruning weights has relatively minimal impact on accuracy and may even yield improvements, particularly on large-scale datasets (\textit{e.g.},  $1.8\%$ ROC-AUC increase on OGBG-HIV with $50\%$ sparsity). \textbf{5)} After halving the number of network layers on three datasets, the performance only drops by $0.2\%–3.3\%$, with the number of parameters decreasing by $32.9\%–45.6\%$ and FLOPs decreasing by $46.6\%–48.8\%$. \textbf{6)} Finally, the extent of accuracy drop depends on the original network size: at the same scale of FLOPs reduction, smaller networks (\textit{e.g.}, GraphTrans on NCI1) exhibit a bigger accuracy drop than large networks (\textit{e.g.}, GraphTrans on OGBG-HIV).



\begin{figure}[!t] 
\begin{center}
\includegraphics[width=1.0\linewidth]{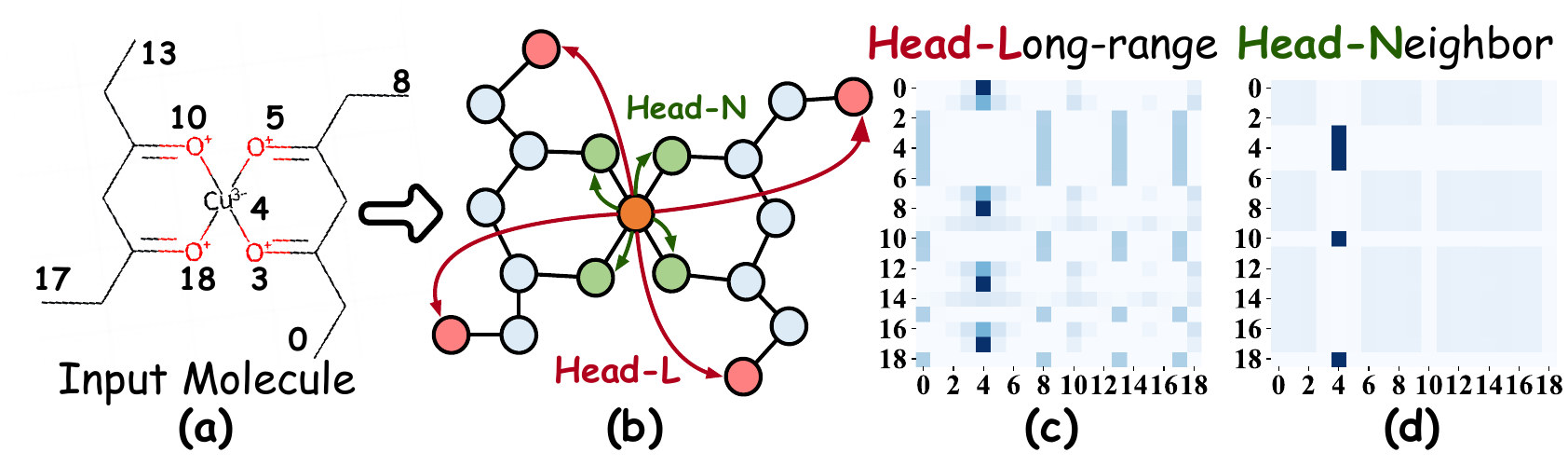}
\end{center}
\caption{Characterizing the roles of different attention heads. Attention heatmaps (subfigures c and d) are shown for the molecule from OGBG-HIV.}
\label{fig:ana-head}
\end{figure}


\begin{figure}[!t] 
\begin{center}
\includegraphics[width=0.99\linewidth]{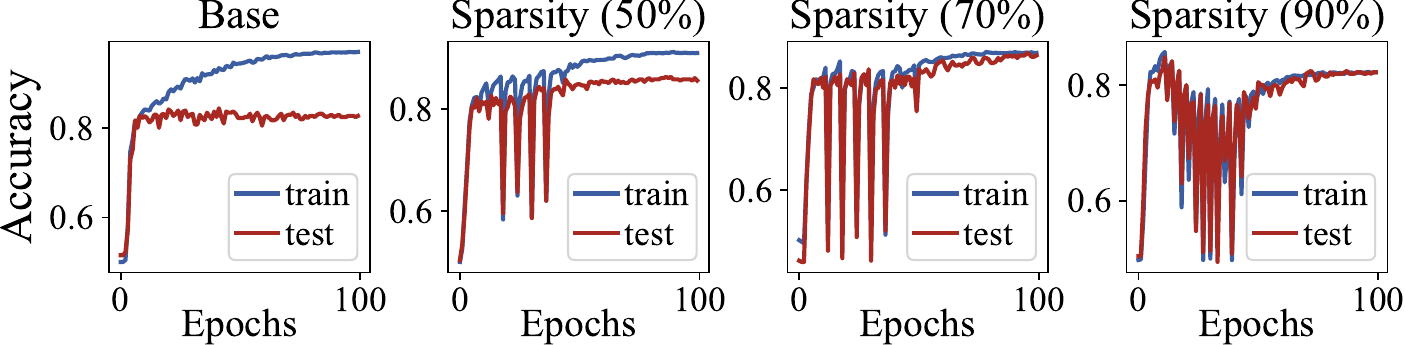}
\end{center}
\caption{Training dynamics \textit{w.r.t.} epochs of dense model (Base) and our GTSP-WP with varying sparsity levels.}
\label{fig:ana-over-fitting}
\end{figure}
\subsection{Generalization to Other GT Models}
\label{sec:general}
To validate the generalizability of our GTSP, we apply it to another two representative GT baseline models: GraphGPS and Graphormer. The results in Table~\ref{tab:other-gt} demonstrate that GTSP effectively compresses these models. Despite a relatively low sparsity ratio, GTSP can match or even exceed the performance of the baseline models. These findings confirm that our GTSP is architecture-independent and can be easily integrated into other GT models.

\subsection{Broader Evaluation of GTSP}
\label{sec:broader}

\paragraph{Unique roles of attention heads} 
We investigate whether heads in the model have interpretable roles. In Figure~\ref{fig:ana-head}, we visualize two attention score matrices from a model trained on OGBG-HIV. It is observed that each head concentrates on distinct nodes:  \textbf{Head-L} captures long-range information by attending to distant nodes, while \textbf{Head-N} focuses on neighboring nodes. These findings indicate that the heads have identifiable functions and are highly interpretable.

\paragraph{Pruning weights help alleviate over-fitting}
In training, over-fitting frequently occurs in GT models due to limited graph data and the large number of parameters. In this section, we will investigate whether our weight pruning strategy effectively alleviates this common problem. Figure~\ref{fig:ana-over-fitting} illustrates the training progress with respect to epochs for the dense model (baseline) and various sparse models with different degrees of sparsity. It is evident that as the sparsity increases, the training curve comes to more closely approximate the test curve. This confirms that our weight pruning strategies do indeed help to alleviate over-fitting in GTs.

\begin{figure}[!t] 
\begin{center}
\includegraphics[width=0.95\linewidth]{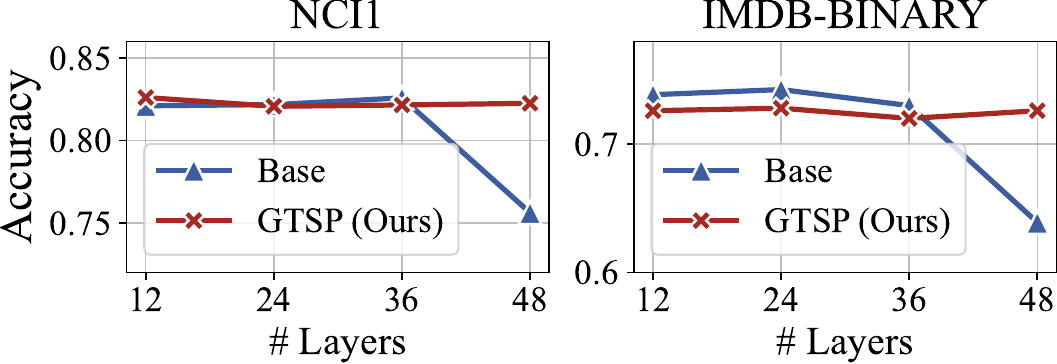}
\end{center}
\caption{Accuracy \textit{w.r.t.} the number of layers on two datasets. Base refers to the GraphTrans model.}
\label{fig:ana-over-smoothing}
\end{figure}

\begin{figure}[!t] 
\begin{center}
\includegraphics[width=0.97\linewidth]{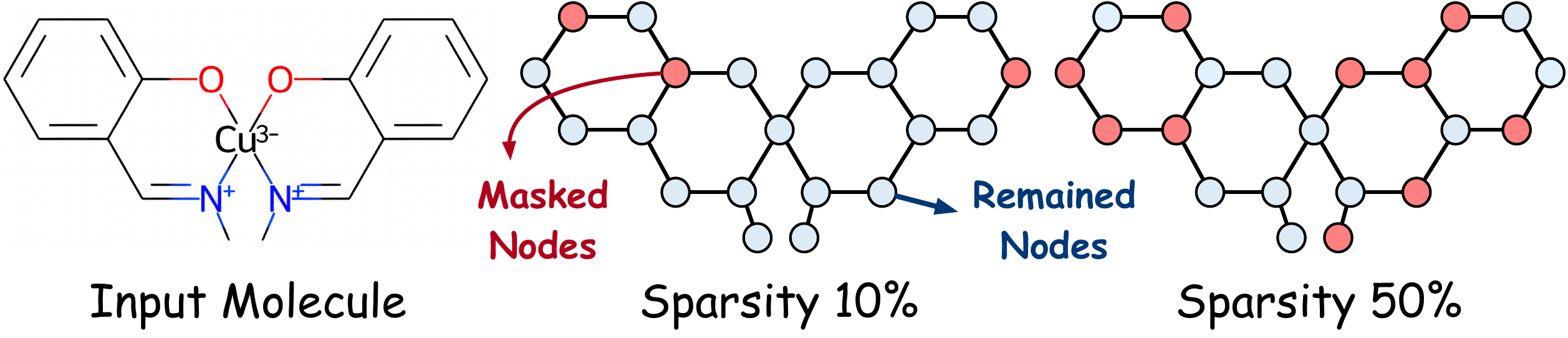}
\end{center}
\caption{Token selection visualization on OGBG-HIV.}
\label{fig:ana-token}
\end{figure}

\paragraph{Pruning layers help alleviate over-smoothing} 
Our previous experiments have confirmed both that the GT model layer contains redundancy and that our pruning strategy effectively reduces it. However, in our previous network, we assumed a network depth of only four layers. Recent research suggests that when the network reaches a certain depth (\textit{e.g.}, 48 layers), it not only leads to redundancy but also causes over-smoothing~\cite{deepgraph}. Therefore, we are conducting further research to investigate whether our pruning strategy can mitigate the over-smoothing problem in excessively deep models. Figure~\ref{fig:ana-over-smoothing} illustrates that while the accuracy of the baseline model drops suddenly at 48 layers, our GTSP maintains performance and significantly outperforms baselines. These experiments demonstrate that our GTSP helps to alleviate over-smoothing and has the potential for breaking the depth limitation of GTs.


\paragraph{Graph Transformer prioritizes informative nodes} 
To further investigate the behavior of GTSP, we present a visualization of the token selection process during testing in Figure~\ref{fig:ana-token}. The results demonstrate that our GTSP primarily focuses on chemical atoms or motifs, rather than other common motifs such as benzene rings. This indicates that our GTSP can effectively distinguish informative nodes from less-informative ones. Moreover, this phenomenon suggests that GTSP enhances the interpretability of GTs by identifying the key nodes in the graph that contribute significantly to graph property identification.


\section{Conclusion}


In this paper, we propose GTSP, a framework that compresses GT models by reducing the redundancy in input data, attention heads, model layers, and model weights. Extensive experiments on large-scale datasets demonstrate the effectiveness and  generalizability of GTSP. Furthermore, the experimental results offer several valuable insights into existing GTs, which can potentially inspire further research. However, there remain several challenges. For instance, adjusting the pruning ratio for different graphs and jointly pruning all components while considering their complicated interactions merit further exploration.

\bibliography{aaai23}

\end{document}